\documentclass[10pt,twocolumn,letterpaper]{article}

\usepackage{cvpr}      

\definecolor{cvprblue}{rgb}{0.21,0.49,0.74}
\usepackage[pagebackref,breaklinks,colorlinks,allcolors=cvprblue]{hyperref}
\usepackage{dblfloatfix}
\usepackage{multirow}

\usepackage{threeparttable}

\newcommand\Tstrut{\rule{0pt}{2.2ex}}         

\def\paperID{88} 
\def\confName{EarthVision}
\def\confYear{2025}

\title{Hyperspectral Vision Transformers for Greenhouse Gas Estimations from Space}

\author{
Ruben Gonzalez Avilés$^{1,2}$ \quad
Linus Scheibenreif$^1$ \quad
Nassim Ait Ali Braham$^3$ \quad
Benedikt Blumenstiel$^2$ \quad
\and
Thomas Brunschwiler$^2$ \quad
Ranjini Guruprasad$^2$ \quad
Damian Borth$^1$ \quad
Conrad Albrecht$^3$ \quad
\and
Paolo Fraccaro$^2$ \quad
Devyani Lambhate$^2$ \quad
Johannes Jakubik$^2$
\\
$^1$\textbf{University of St. Gallen}
$^2$\textbf{IBM Research}
$^3$\textbf{German Aerospace Center (DLR)}
}

\begin{document}
\maketitle
\begin{abstract}

Hyperspectral imaging provides detailed spectral information and holds significant potential for monitoring of greenhouse gases (GHGs). However, its application is constrained by limited spatial coverage and infrequent revisit times. In contrast, multispectral imaging offers broader spatial and temporal coverage but often lacks the spectral detail that can enhance GHG detection. To address these challenges, this study proposes a spectral transformer model that synthesizes hyperspectral data from multispectral inputs. The model is pre-trained via a band-wise masked autoencoder and subsequently fine-tuned on spatio-temporally aligned multispectral–hyperspectral image pairs. The resulting synthetic hyperspectral data retain the spatial and temporal benefits of multispectral imagery and improve GHG prediction accuracy relative to using multispectral data alone. This approach effectively bridges the trade-off between spectral resolution and coverage, highlighting its potential to advance atmospheric monitoring by combining the strengths of hyperspectral and multispectral systems with self-supervised deep learning.

\end{abstract}    
\section{Introduction}
\label{sec:intro}

Satellite-based remote sensing is vital to a wide range of Earth observation applications, from land-use mapping and agricultural assessments to climate and atmospheric studies \cite{lary2016machine}. By measuring reflected or emitted radiation across multiple wavelengths, remote sensing provides large-scale datasets for analyzing environmental changes, land cover characteristics, and other geophysical phenomena \cite{miller-2024}. Among satellite imaging systems, hyperspectral sensors (e.g., EnMAP) capture hundreds of narrow, contiguous bands, enabling fine-grained spectral analysis \cite{9324006, bue-2010}. However, these systems often have narrower swath widths and longer revisit intervals \cite{zhang-2016}, limiting their spatial and temporal coverage.

Multispectral sensors (e.g., Sentinel-2, Landsat) instead collect fewer, broader bands but provide wider coverage and more frequent revisits \cite{sentinel-2, landsat-nasa}. This makes multispectral imagery well suited for large-scale, time-sensitive applications, but its coarser spectral resolution may overlook subtle absorption features, such as those associated with trace-gases like carbon dioxide (CO\textsubscript{2}), methane (CH\textsubscript{4}), or nitrogen dioxide (NO\textsubscript{2}). A technique that fuses the richer spectral detail of hyperspectral imagery with the extensive coverage of multispectral sensors could yield more comprehensive remote sensing capabilities \cite{zhang-2016}.

\section{Background}
\label{sec:background}

\subsection{Multispectral and Hyperspectral Imaging}

Remote sensing acquires information about the Earth’s surface and atmosphere by measuring electromagnetic radiation reflected or emitted by various materials \cite{aggarwal2004principles}. While both multispectral and hyperspectral imaging exploit this principle, they differ in spectral resolution and coverage. Multispectral systems observe a few broad bands (e.g., Sentinel-2 has 13), providing rapid, wide-scale coverage suitable for tasks like land cover classification and vegetation monitoring \cite{sentinel-2, Kogan2019}. Their broader bands, however, may miss subtle absorption features that can be important for detailed material or gas analyses \cite{groh-2005}.

\begin{figure}[!h]
    \centering
    \includegraphics[width=\columnwidth]{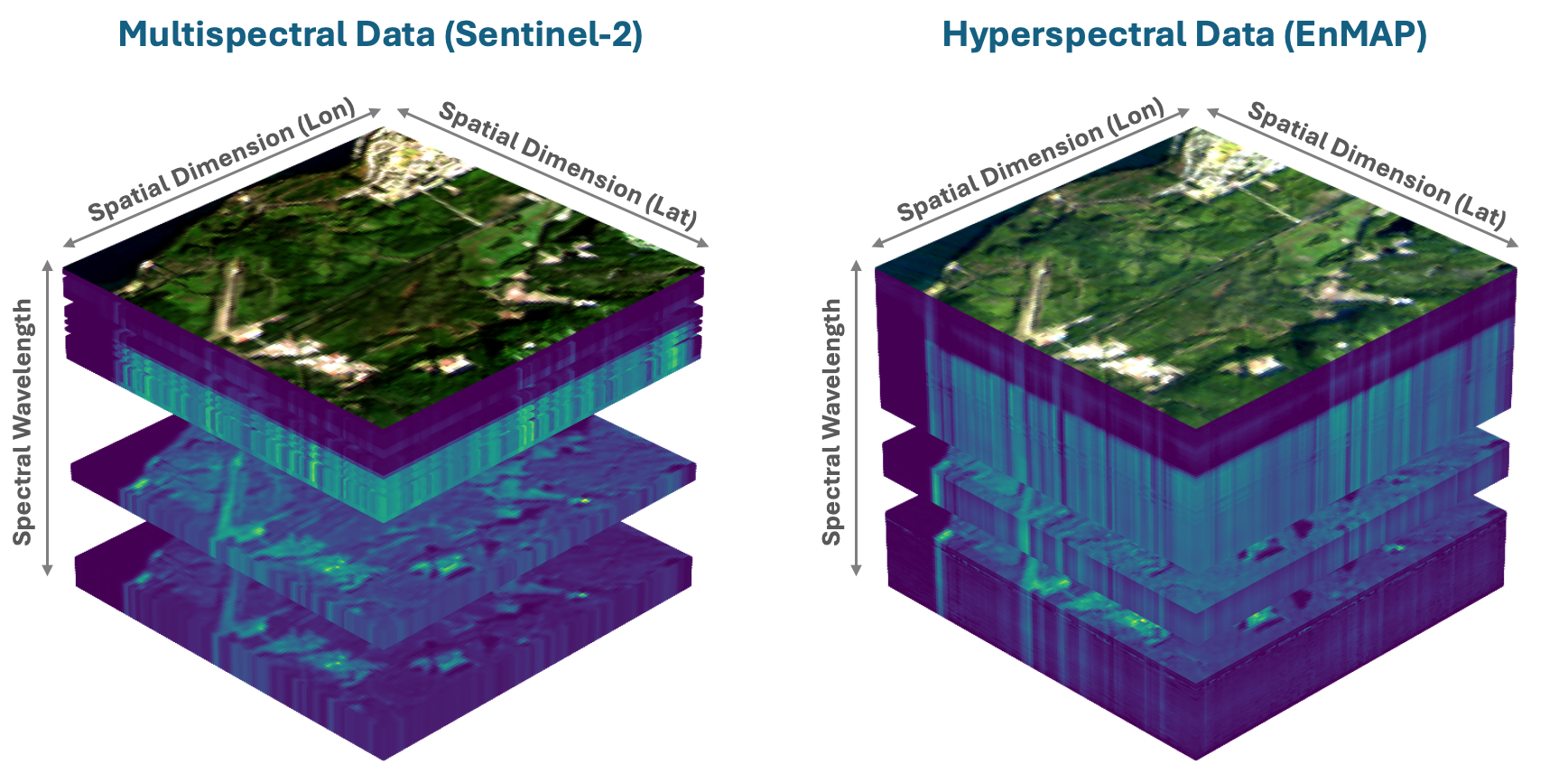}
    \caption{3D visualization of corresponding Sentinel-2 (left) and EnMAP (right) data cubes, each spanning the same spatial dimensions along the horizontal axes and the spectral dimension along the vertical axis.}
    \label{fig:msi_vs_hsi}
\end{figure}

\begin{figure*}[!hb]
  \centering
    \includegraphics[width=\textwidth]{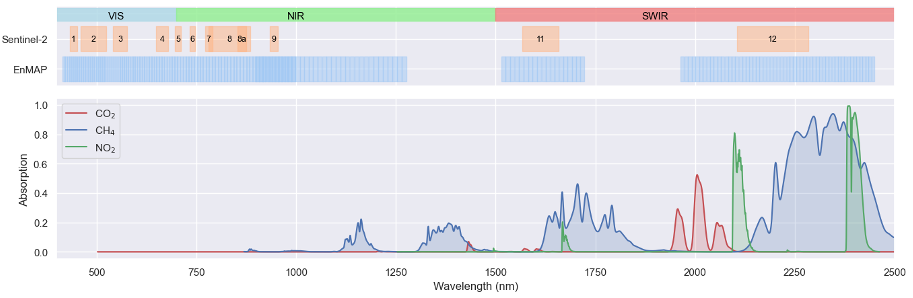}
    \caption{Absorption spectra and satellite band coverage. The upper section illustrates the wavelength coverage of each spectral band, with widths corresponding to the Full-Width Half-Maximum (FWHM) values extracted from the metadata of the corresponding satellite product \cite{sentinel-2, 9324006}. The lower section shows the smoothed absorption spectra for three gases, with absorption data sourced from the High Resolution Transmission (HITRAN) database and smoothed using a Gaussian filter to reduce noise for visualization \cite{gordon-2021}.}
    \label{fig:absorption_spectrum}
\end{figure*}

Hyperspectral systems (e.g., EnMAP) capture hundreds of narrow, contiguous bands that can distinguish fine spectral differences among materials \cite{9324006, shokati-2024}. This facilitates tasks such as precise mineral exploration, species-level vegetation mapping, and gas detection, where subtle wavelength-specific absorption or reflectance patterns can be particularly relevant \cite{danilov-2023, masin-2023}. Yet hyperspectral instruments often have reduced spatial swath widths and less frequent revisit times, creating a trade-off between spectral resolution and the ability to observe large areas or rapidly evolving conditions \cite{zhang-2016, jia-2021}.

\subsection{Spectral Signatures and Gas Detection}

An important aspect of remote sensing is that each material or gas exhibits a characteristic reflectance or absorption pattern across specific wavelengths, known as its spectral signature \cite{10470851}. Greenhouse gases (GHGs) such as CO\textsubscript{2} have narrow absorption peaks in the near-infrared (NIR) and shortwave infrared (SWIR) regions \cite{varchand-2022, gordon-2021}. Figure~\ref{fig:absorption_spectrum} illustrates these absorption features for several GHGs, along with the wavelength coverage of Sentinel-2 and EnMAP. Because multispectral sensors capture fewer, broader bands, they may only partially record these absorption patterns, potentially reducing gas retrieval accuracy. In contrast, hyperspectral sensors typically provide more detailed coverage of these wavelengths, which can facilitate improved detection and quantification of atmospheric gases \cite{schodlok-2022}.

By reconstructing hyperspectral-like information from more frequently available multispectral data, one can potentially identify these absorption features more accurately on a broader temporal and spatial scale. This can extend the utility of multispectral imagery for atmospheric studies, including GHG monitoring, where the finer spectral resolution traditionally found only in hyperspectral data may significantly enhance detection accuracy \cite{alsalem-2023, zimmerman-2023}.

\section{Datasets}
This study employs several satellite-based datasets to develop and evaluate a hyperspectral reconstruction model and its downstream application in GHG prediction. Specifically, two large EnMAP hyperspectral collections are used for training and validation, along with both real and synthetic multispectral Sentinel-2 data. Additional GHG measurements from Sentinel-5P \cite{esa_sentinel5p} and the Orbiting Carbon Observatory 3 (OCO-3) \cite{amt-16-3173-2023} are incorporated to assess the impact of spectral detail on gas concentration estimates.

\subsection{Hyperspectral EnMAP Data}

Two EnMAP-derived datasets were used: HySpecNet-11k \cite{fuchs2023hyspecnet11klargescalehyperspectraldataset} and SpectralEarth \cite{braham2024spectralearthtraininghyperspectralfoundation}. Both provide Level-2A (surface reflectance) images at 30\,m resolution, spanning 202 usable bands from 420 to 2450\,nm. A cloud-filtering step was applied at the scene level, which may still leave individual patches with residual cloud cover exceeding 10\% \cite{fuchs2023hyspecnet11klargescalehyperspectraldataset}.

\begin{itemize}
    \item \textbf{HySpecNet-11k:} Comprises 11{,}483 hyperspectral images, each sized at 128$\times$128 pixels, sampled from 250 distinct EnMAP tiles captured between November~2 and November~9, 2022 \cite{fuchs2023hyspecnet11klargescalehyperspectraldataset}.
    \item \textbf{SpectralEarth:} Contains over 538{,}000 128$\times$128-pixel hyperspectral images, derived from more than 11{,}600 EnMAP tiles spanning April~2022 to April~2024 \cite{braham2024spectralearthtraininghyperspectralfoundation}.
\end{itemize}

\subsection{Multispectral Sentinel-2 Data}

To provide multispectral inputs matched to the HySpecNet-11k dataset, we acquired Sentinel-2 Level-2A (surface reflectance) scenes from Google Earth Engine \cite{esa_sentinel2_sr_harmonized}. For each HySpecNet-11k patch, we identified all Sentinel-2 scenes within a \(\pm3\)-day window, then filtered them to \(\leq10\%\) cloud cover over the region of interest. From this filtered set, the scene captured closest in time to the EnMAP acquisition date was selected. This process yielded 9,797 valid EnMAP--Sentinel-2 pairs, covering approximately 85.32\% of the HySpecNet-11k dataset.

\paragraph{Synthetic Sentinel-2 Data}
Because the EnMAP and Sentinel-2 scenes are acquired at slightly different times, they can exhibit varying cloud occlusion patterns, which in turn lead to spatial inconsistencies, especially when residual clouds are present in one image but not in the other. To facilitate controlled comparisons of spectral resolution impacts, a \emph{Synthetic Sentinel-2} dataset was derived by degrading EnMAP hyperspectral data using Sentinel-2A's spectral response function (SRF). This ensures that the only distinction between the Synthetic Sentinel-2 and EnMAP data is the spectral granularity and coverage, eliminating potential confounding factors such as sensor characteristics, processing pipelines, cloud occlusion or temporal mismatches \cite{li-2023, zheng-2021}.

The Synthetic Sentinel-2 data was constructed by mapping EnMAP's spectral bands (\( B_{\text{EnMAP}} \)) to Sentinel-2's (\( B_{\text{S2}} \)) using the SRF. For each Sentinel-2 band \( b \), a weight vector \( \mathbf{w}_b \) was computed based on the SRF values overlapping the wavelength range of each EnMAP band \(i\). The weights were normalized to ensure their sum equaled 1:

\begin{equation}
w_{i,b} = \sum_{\lambda \in [\lambda_{i}^{\min}, \lambda_{i}^{\max}]} \text{SRF}_b(\lambda), \quad \mathbf{\tilde{w}}_b = \frac{\mathbf{w}_b}{\mathbf{1}^\top \mathbf{w}_b}.
\label{eq:normalized_weights}
\end{equation}

The complete weight matrix \( \mathbf{W} = \left[\mathbf{\tilde{w}}_1,\, \dots,\, \mathbf{\tilde{w}}_{B_{\text{S2}}}\right] \in \mathbb{R}^{B_{\text{EnMAP}} \times B_{\text{S2}}} \) was then applied to project the hyperspectral data \( \mathbf{X} \in \mathbb{R}^{H \times W \times B_{\text{EnMAP}}} \) into the Sentinel-2 spectral space:

\begin{equation}
\mathbf{Y}_{h,w,:} = \mathbf{X}_{h,w,:} \cdot \mathbf{W}, \quad \forall \, h, w.
\label{eq:ss2_projection}
\end{equation}

Here, \( \mathbf{Y} \in \mathbb{R}^{H \times W \times B_{\text{S2}}} \) represents the resulting Synthetic Sentinel-2 dataset, capturing Sentinel-2's spectral characteristics while preserving the spatial and temporal properties of EnMAP data.

\subsection{Greenhouse Gas Data}

We assessed the reconstructed hyperspectral data's utility for predicting concentrations of NO\textsubscript{2}, CH\textsubscript{4}, and CO\textsubscript{2}. These GHG products were spatiotemporally mapped to the HySpecNet-11k dataset based on the spatial footprint of EnMAP patches under application-specific time windows:

\begin{itemize}
    \item \textbf{Sentinel-5P \cite{esa_sentinel5p}:}
    \begin{itemize}
        \item \textbf{NO\textsubscript{2}} retrievals (\(\mathrm{mol\,m^{-2}}\)) were matched within \(\pm 1\) day, reflecting its shorter atmospheric lifetime~\cite{lange-2022}. This yielded data for about 88\% of the EnMAP patches.
        \item \textbf{CH\textsubscript{4}} measurements (ppb) were matched within \(\pm 7\) days, consistent with methane's longer lifetime~\cite{saunois-2016}. Approximately 52\% of EnMAP patches were matched with valid CH\textsubscript{4} values.
    \end{itemize}

    \item \textbf{OCO-3 \cite{amt-16-3173-2023} (CO\textsubscript{2}):} Measurements (ppm) were sourced from OCO-3, an instrument aboard the International Space Station (ISS), using a \(\pm 16\)~day window to account for CO\textsubscript{2}'s extended atmospheric residence time \cite{archer-2009}. Only 5.5\% of EnMAP patches had suitable OCO-3 coverage, owing to the spatially limited extent of OCO-3 soundings and the resulting sparse spatiotemporal overlap with EnMAP footprints.
\end{itemize}
\section{Methods}
\label{sec:methods}

\subsection{Spectral Transformer Masked Autoencoder}
\label{sec:spectral_transformer_mae}

\begin{figure*}[!ht]
  \centering
    \includegraphics[width=\textwidth]{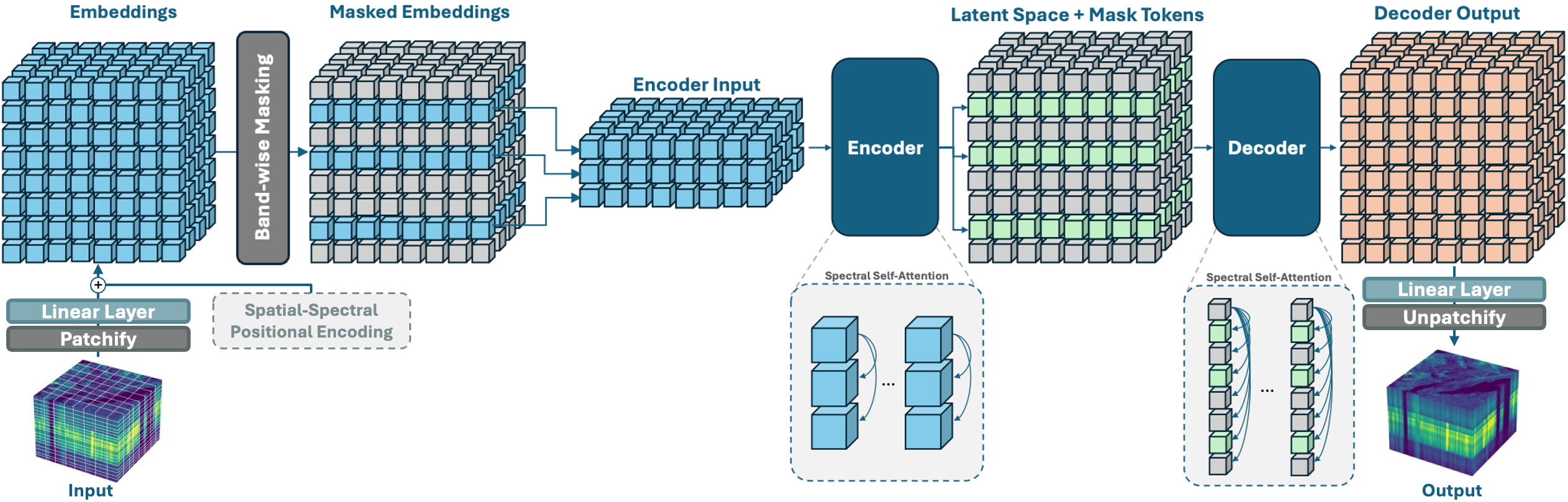}
    \caption{High-level overview of the masked autoencoder approach for hyperspectral image reconstruction. The model applies band-wise masking to input patches, encodes the latent representations using spectral self-attention, and reconstructs the missing spectral information through a decoder.}
    \label{fig:model_overview}
\end{figure*}

We introduce a spectral transformer model designed to reconstruct hyperspectral data from limited spectral inputs using a self-supervised masked autoencoder framework, as depicted in Figure~\ref{fig:model_overview}. Our model leverages a spectral self-attention mechanism to capture inter-band dependencies at each spatial location. Specifically, for each spatial position, tokens corresponding to different spectral bands attend to one another. In the encoder, spectral attention is applied solely to the visible (unmasked) tokens, allowing the network to extract robust latent representations of the available spectral information. In the decoder, however, attention is computed over both visible and masked tokens, enabling the reconstruction of missing spectral bands. Finally, a linear layer projects the decoder outputs back into pixel space, and the patch-level reconstructions are rearranged to form a complete hyperspectral data cube that preserves both spectral fidelity and spatial coherence. Finally, the reconstructed hyperspectral data cube is compared to the original full hyperspectral cube using the Mean Absolute Error (MAE) objective function.

\paragraph{Preprocessing and Patch Embedding}
We apply band-wise normalization using means \(\mu_b\) and standard deviations \(\sigma_b\) computed over the training dataset with Welford's algorithm~\cite{welford-1962}:
\begin{equation}
\tilde{x}_{b,:,:} \;=\; \frac{x_{b,:,:} - \mu_b}{\sigma_b}, \quad \forall\, b \in \{1,\dots,B\},
\label{eq:bandwisenorm}
\end{equation}
ensuring each spectral band is on a comparable scale. The normalized hyperspectral cube 
\(\tilde{X} \in \mathbb{R}^{B \times H \times W}\) (with \(B\) bands and spatial dimensions \(H \times W\)) 
is then divided into patches of size \(b \times p \times p\), where \(b\) is a subset of bands and \(p\) specifies the spatial patch dimensions.

Let \(P_i \in \mathbb{R}^{b \times p \times p}\) be the \(i\)-th patch. We flatten \(P_i\) into 
\(\text{Flatten}(P_i) \in \mathbb{R}^{b\,p^2}\), then map it into a fixed-dimensional embedding 
\(\mathbf{E}_i \in \mathbb{R}^d\) via a learnable linear projection:
\begin{equation}
\mathbf{E}_i \;=\; \mathbf{W}_{\mathrm{embed}} \,\text{Flatten}(P_i) \,+\, \mathbf{b}_{\mathrm{embed}},
\label{eq:linearlayer}
\end{equation}
where \(\mathbf{W}_{\mathrm{embed}} \in \mathbb{R}^{d \times (b\,p^2)}\) and 
\(\mathbf{b}_{\mathrm{embed}} \in \mathbb{R}^d\). 
This embedding preserves both spectral and spatial information, preparing the patches for subsequent model processing.

\paragraph{Spatial-Spectral Positional Encoding}
We enrich each patch embedding with a 2D spatial encoding and a 1D spectral encoding based on the patch’s grid coordinates \((x,y)\) and its center wavelength \(\lambda\). For the spatial dimension, we use sinusoidal functions. With an embedding dimension \(d\), for each index \(i\) we define:
\begin{align}
E_{\text{spatial}}(x,2i) &= \sin\!\Bigl(\frac{x}{10000^{\,2i/d}}\Bigr), \\
E_{\text{spatial}}(x,2i+1) &= \cos\!\Bigl(\frac{x}{10000^{\,2i/d}}\Bigr),
\end{align}
and similarly for \(y\). 

For the spectral dimension, we first normalize the center wavelength \(\lambda\) into the range \([0,1]\) and then scale it to align with the magnitude of the spatial encoding. This is done as follows:
\begin{equation}
\lambda_{\text{scaled}} = \frac{\lambda - \lambda_{\min}}{\lambda_{\max} - \lambda_{\min}} \cdot N_{\text{spatial}},
\label{eq:wavelengthnorm}
\end{equation}
where \(\lambda_{\min}=400\,\mathrm{nm}\) and \(\lambda_{\max}=2500\,\mathrm{nm}\) are the lower and upper bounds of the spectral range, and \(N_{\text{spatial}}\) is the number of spatial indices (e.g., the grid size of patches in one dimension) used for scaling. The scaled wavelength is then encoded using sinusoidal functions:
\begin{align}
E_{\text{spectral}}(\lambda,2i) &= \sin\!\Bigl(\frac{\lambda_{\text{scaled}}}{10000^{\,2i/d}}\Bigr), \\
E_{\text{spectral}}(\lambda,2i+1) &= \cos\!\Bigl(\frac{\lambda_{\text{scaled}}}{10000^{\,2i/d}}\Bigr).
\end{align}

The final positional encoding is obtained by summing the spatial and spectral encodings element-wise:
\begin{equation}
E_{\text{pos}}(x,y,\lambda) = E_{\text{spatial}}(x,y) + E_{\text{spectral}}(\lambda).
\label{eq:posenc}
\end{equation}
This combined encoding is added to the patch embedding \(E_i\) to yield the initial token:
\begin{equation}
Z_i^0 = E_i + E_{\text{pos}}(x_i,y_i,\lambda_i).
\label{eq:posencapplied}
\end{equation}
This process enables the model to incorporate both the spatial location and the spectral center of each patch into its latent representation.

\paragraph{Band-wise Masking Strategy}
To train the model in a self-supervised manner, a fraction \(p_{\text{mask}}\) of the spectral indices is randomly selected for masking. Let \(M \subset \{1,\dots,B\}\) denote the set of masked spectral indices, where \(|M| = p_{\text{mask}} \times B\). For each token \(Z_i^0\) corresponding to spectral index \(i\), we define the resulting token as:
\begin{equation}
\tilde{Z}_i^0 \;=\;
\begin{cases}
Z_{\text{mask}}, & \text{if } i \in M, \\
Z_i^0, & \text{otherwise,}
\end{cases}
\label{eq:bandwisemasking}
\end{equation}
where \(Z_{\text{mask}}\) is a learnable mask token that signals the absence of information, enabling the model to encode missing data and reconstruct it from the context provided by unmasked tokens.

\subsection{Greenhouse Gas Estimation Model}
\label{sec:ghg_model}

To assess the effectiveness of our reconstructed hyperspectral data, we evaluate its performance in a downstream greenhouse gas estimation task. As shown in Fig.~\ref{fig:absorption_spectrum}, the absorption features are more comprehensively captured by the dense spectral coverage of EnMAP compared to the sparser bands in multispectral imagery. We hypothesize that accurately reconstructing these finer details from multispectral inputs will yield improved GHG predictions relative to the multispectral baseline. It is important to note that our goal is not to maximize absolute GHG accuracy, but rather to compare the efficacy of different input spectra (original EnMAP, reconstructed hyperspectral, and Sentinel-2 data) in capturing critical spectral information.

\paragraph{Input Preparation}
To obtain a representative spectral signature for each image patch, we compute the mean reflectance across all spatial pixels. Formally, for each spectral band \(b\), the signature is given by
\begin{equation}
\mathbf{x}_b = \frac{1}{N} \sum_{i=1}^{N} \mathbf{R}_{b,i},
\label{eq:spatialaveraging}
\end{equation}
where \(\mathbf{R}_{b,i}\) denotes the reflectance at pixel \(i\) for band \(b\), and \(N\) is the total number of spatial pixels in the patch. This averaging process aligns with the coarser spatial resolution of GHG measurements (e.g., Sentinel-5P, OCO-3) and emphasizes the spectral characteristics of each patch while removing any spatial features.

\paragraph{Data Normalization}
For the downstream regression, we apply a global normalization:
\begin{equation}
\tilde{\mathbf{x}} = \frac{\mathbf{x} - \mu}{\sigma},
\label{eq:globalnorm}
\end{equation}
where \(\mu\) and \(\sigma\) are the mean and standard deviation computed across \emph{all} bands and training samples. This retains relative inter-band differences that can be relevant for gas detection, in contrast to the band-wise normalization used during hyperspectral reconstruction.

\paragraph{Model Architecture}
A simple feedforward network (multilayer perceptron, MLP) regresses from the spectral signature to a single GHG concentration. The input size depends on whether the data is (i) Sentinel-2 (12 bands), (ii) original EnMAP (202 bands), or (iii) reconstructed EnMAP-like spectra. We use mean squared error (MSE) loss to penalize large deviations more heavily, thus emphasizing the fidelity of reconstructed bands.

\section{Experimental Setup}
\label{sec:ex_setup}

\begin{figure}[!h]
    \centering
    \includegraphics[width=\columnwidth]{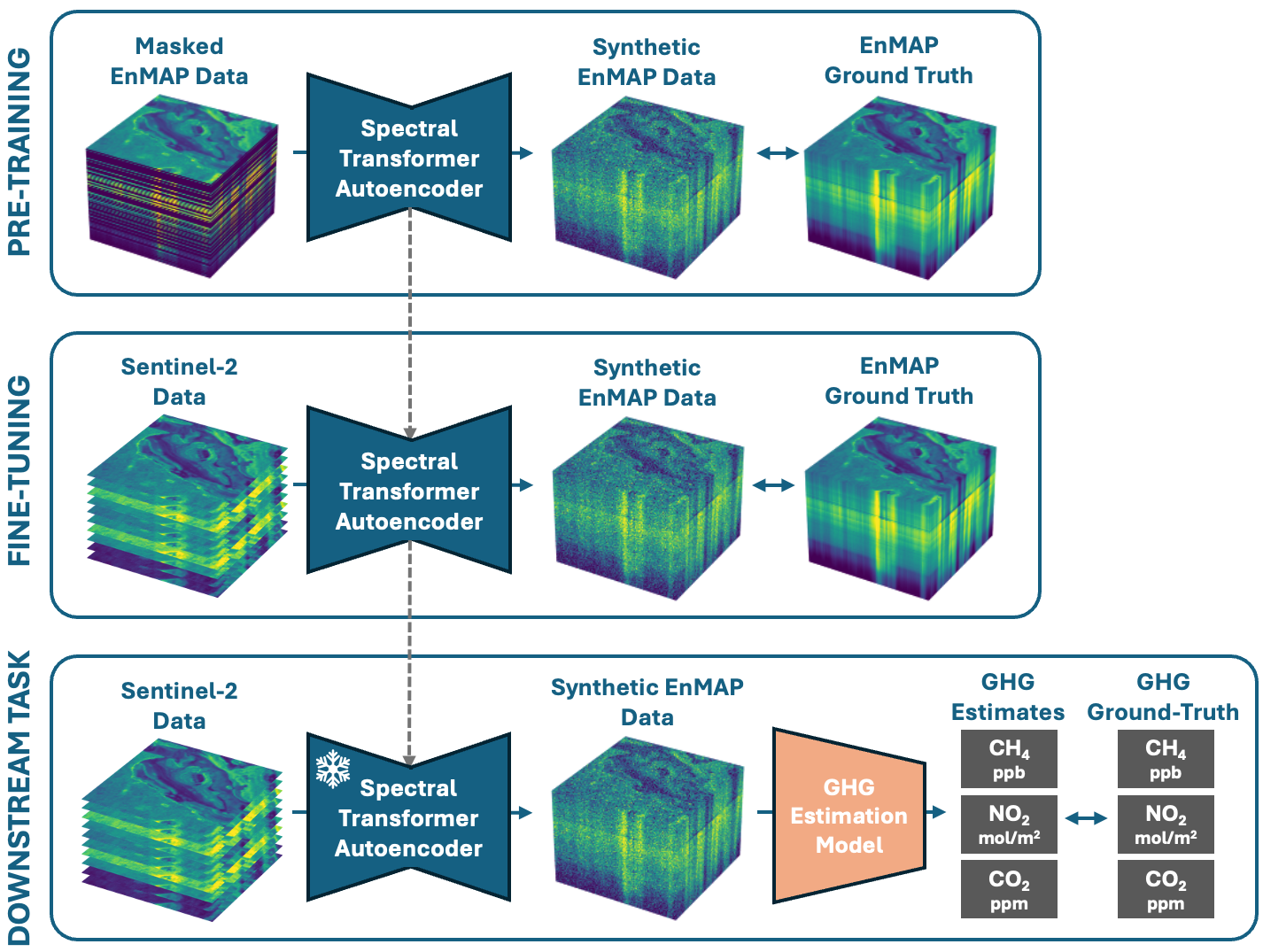}
    \caption{High-level overview of the three-stage experimental setup. The model is first pre-trained on masked hyperspectral data, then fine-tuned on multispectral inputs to reconstruct hyperspectral-like data, and finally evaluated on a downstream GHG estimation task.}
    \label{fig:experimental_setup}
\end{figure}

We evaluated the spectral transformer model using a three-stage framework, as illustrated in Fig.~\ref{fig:experimental_setup}. In the first stage, the model is pre-trained on masked hyperspectral inputs. In the second stage, the pre-trained model is fine-tuned using multispectral inputs to reconstruct hyperspectral images. Finally, the reconstructed hyperspectral data are applied to a downstream task for GHG estimation.

\subsection{Dataset Splits}
\label{sec:dataset_splits}

The HySpecNet-11k dataset provides two predefined splits: an \emph{easy} split, where images are randomly partitioned into training, validation, and test sets, and a \emph{hard} split, where it is ensured that images in different sets are sourced from distinct EnMAP tiles \cite{fuchs2023hyspecnet11klargescalehyperspectraldataset}.

\paragraph{Reconstruction Task}
For the reconstruction task, we used the \emph{hard} split to minimize potential information leakage between training, validation, and test sets. This strict separation ensures that the model is evaluated on entirely unseen EnMAP tiles. Moreover, when pre-training on the larger SpectralEarth dataset, any overlapping tiles with the HySpecNet-11k test set were excluded from the training set.

\paragraph{Downstream GHG Task}
For greenhouse gas estimation, we employed the \emph{easy} split, which randomly partitions the images. Since the ground-truth labels (e.g., from Sentinel-5P or OCO-3) are unique to each patch, this split maintains a diverse geographical distribution while minimizing label duplication. All models in this task were trained and evaluated under identical conditions for fair comparison.

\subsection{Reconstruction Task}
\label{sec:recon_task}

\paragraph{Pre-Training on Hyperspectral Data}
The spectral transformer autoencoder model was pre-trained on masked hyperspectral data from the HySpecNet-11k and SpectralEarth datasets using a masking fraction of 80\%. This pre-training enables the model to learn spectral dependencies across the entire wavelength range, establishing a foundation that can be fine-tuned on satellite products with sparse spectral coverage regardless of the specific wavelengths or number of bands available.

\paragraph{Fine-Tuning on Multispectral Data}
After pre-training, we replaced the masked hyperspectral inputs with multispectral data (true Sentinel-2 and synthetic Sentinel-2) mapped to the same EnMAP patches. The model was then fine-tuned to reconstruct the full hyperspectral range from these coarser inputs, with performance evaluated by comparing the reconstructed cubes to the original EnMAP data on the test set.

\subsection{Downstream GHG Task}
\label{sec:downstream_task}

We next assessed whether reconstructed hyperspectral data improves GHG estimation. Each EnMAP patch (or its multispectral/reconstructed counterpart) was spatially averaged to a single spectral signature, then fed into a feedforward regressor predicting NO\textsubscript{2}, CH\textsubscript{4}, or CO\textsubscript{2} concentrations. We trained separate models using:
\begin{enumerate}[label=(\roman*)]
    \item \textit{Original EnMAP (hyperspectral) data} -- upper baseline;
    \item \textit{Sentinel-2 (multispectral) data} -- lower baseline;
    \item \textit{Reconstructed hyperspectral} from Sentinel-2 inputs -- proposed approach.
\end{enumerate}
Improvements over the multispectral baseline indicate that the reconstruction retains valuable spectral information.

\subsection{Evaluation Metrics}
\label{sec:evaluation_metrics}

For hyperspectral reconstruction, we compute the following metrics. The Mean Absolute Error (MAE) is defined as
\begin{equation}
\text{MAE}(X,\hat{X}) = \frac{1}{N} \sum_{i=1}^{N} \left| X_i - \hat{X}_i \right|,
\label{eq:mae}
\end{equation}
where \(X\) and \(\hat{X}\) denote the ground truth and reconstructed data, respectively. The Peak Signal-to-Noise Ratio (PSNR) is computed as
\begin{equation}
\text{PSNR}(X,\hat{X}) = 10 \log_{10}\!\left(\frac{\max(X)^2}{\text{MSE}(X,\hat{X})}\right),
\label{eq:psnr}
\end{equation}
with the Mean Squared Error (MSE) given by
\begin{equation}
\text{MSE}(X,\hat{X}) = \frac{1}{N}\sum_{i=1}^{N}\left( X_i - \hat{X}_i \right)^2.
\label{eq:mse}
\end{equation}
The Structural Similarity Index (SSIM) is computed using the default values from the \textit{torchmetrics} library (i.e., \(C_1=0.01\) and \(C_2=0.03\)) \cite{torchmetrics_ssim}:
\begin{equation}
\text{SSIM}(X,\hat{X}) = \frac{(2\mu_X\mu_{\hat{X}}+0.01)(2\sigma_{X\hat{X}}+0.03)}{(\mu_X^2+\mu_{\hat{X}}^2+0.01)(\sigma_X^2+\sigma_{\hat{X}}^2+0.03)}.
\label{eq:ssim}
\end{equation}
The Spectral Angle Mapper (SAM), which measures the angular difference between spectral signatures, is defined as
\begin{equation}
\text{SAM}(X,\hat{X}) = \arccos\!\left(\frac{\sum_{b=1}^{B} X_b\,\hat{X}_b}{\sqrt{\sum_{b=1}^{B} X_b^2}\,\sqrt{\sum_{b=1}^{B} \hat{X}_b^2}}\right).
\label{eq:sam}
\end{equation}

For downstream greenhouse gas prediction, we use the same definitions for MAE and MSE (with \(y\) and \(\hat{y}\) denoting ground-truth and predicted gas concentrations) and additionally compute the Root Mean Squared Error (RMSE) as
\begin{equation}
\text{RMSE}(y,\hat{y}) = \sqrt{\text{MSE}(y,\hat{y})},
\label{eq:rmse}
\end{equation}
and the \(R^2\) Score, which quantifies the fraction of variance explained by the model, as
\begin{equation}
R^2 = 1 - \frac{\sum_{i=1}^{N} (y_i - \hat{y}_i)^2}{\sum_{i=1}^{N} (y_i - \bar{y})^2}.
\label{eq:r2}
\end{equation}
These metrics together provide a comprehensive assessment of both the reconstruction fidelity and the practical utility of the reconstructed data for greenhouse gas detection.

\section{Experimental Results}
\label{sec:ex_results}

\subsection{Reconstruction Results}
\label{sec:recon_results}

We first evaluated the model’s ability to reconstruct hyperspectral imagery from two input types: (i) masked EnMAP data (self-supervised pre-training) and (ii) multispectral Sentinel-2 data (fine-tuning). Our goal is to assess how effectively the model reconstructs the complete hyperspectral data cube when substantial portions of the spectral range are missing or when only multispectral data is provided.

\paragraph{Pre-training on Masked EnMAP Data}

Pre-training the spectral transformer autoencoder on masked EnMAP data using the HySpecNet-11k and the larger SpectralEarth dataset yielded reconstruction results with lower MAE, higher PSNR and SSIM, and reduced SAM compared to simpler approaches such as Gaussian Sampling and Linear Interpolation, as summarized in Table~\ref{tab:masked_enmap}.

\begin{table}[h!]
\centering
\small
\caption{Quantitative reconstruction results from masked EnMAP data.}
\label{tab:masked_enmap}
\begin{tabular}{lcccc}
\hline
\textbf{Model}  & \textbf{MAE↓} & \textbf{PSNR↑} & \textbf{SSIM↑} & \textbf{SAM↓} \\
\hline
Gaussian Sampling      & 848.83 & 24.65 & 0.532 & 39.44 \\
Linear Interpolation   & 139.98 & 36.40 & 0.971 & 12.13 \\
\hline
Masked Autoenc. (HS)   & 41.39  & 41.82 & 0.971 & \textbf{3.68} \\
Masked Autoenc. (SE)   & \textbf{31.82}  & \textbf{42.26} & \textbf{0.973} & 6.46 \\
\hline
\end{tabular}
\begin{tablenotes}
\small
\item \centering \textbf{HS}: HySpecNet-11k; \textbf{SE}: SpectralEarth.
\end{tablenotes}
\end{table}

In our Gaussian Sampling method, values were generated from a Gaussian distribution defined by the band-wise mean and standard deviation, while Linear Interpolation reconstructs masked bands by linearly interpolating from neighboring unmasked bands. These results suggest that the masked autoencoder captures spectral dependencies more effectively than these basic methods. Figure~\ref{fig:masked_reconstruction} provides qualitative examples for selected EnMAP images, where all three RGB bands were fully masked, demonstrating that the reconstructed RGB bands and spectral signatures closely resemble the ground truth.

\begin{figure}[!h]
    \centering
    \includegraphics[width=\columnwidth]{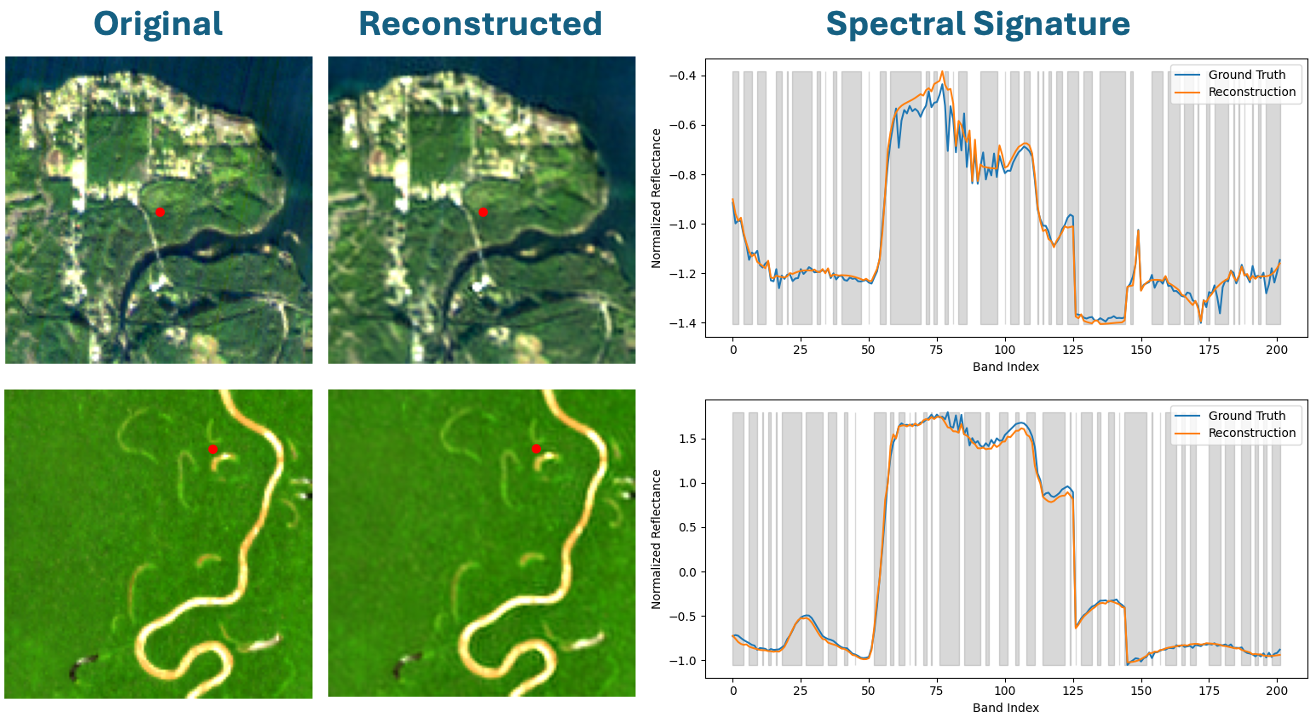}
    \caption{Original and reconstructed RGB bands (left and middle columns) for selected EnMAP images. The right column displays band-wise normalized reflectance at the location of the red dot in each RGB image, with ground-truth signatures in blue and reconstructed signatures in orange. Masked bands are highlighted in gray.}
    \label{fig:masked_reconstruction}
\end{figure}

\paragraph{Fine-tuning on Multispectral Data}
Table~\ref{tab:synth_finetuning} shows that, before fine-tuning, reconstruction performance on multispectral inputs is substantially degraded for both real Sentinel-2 (S2) and synthetic Sentinel-2 (SS2) data. 

\begin{table}[h!]
\centering
\small
\caption{Reconstruction performance comparison for real (S2) and synthetic (SS2) Sentinel-2 data.}
\label{tab:synth_finetuning}
\begin{tabular}{lccccc}
\hline
\textbf{Input} & \textbf{Finetuned} & \textbf{MAE↓} & \textbf{PSNR↑} & \textbf{SSIM↑} & \textbf{SAM↓} \\
\hline
S2  & \(\times\)   & 654.20 & 21.15 & 0.582 & 22.29 \\
S2  & \checkmark   & \textbf{294.14} & \textbf{27.00} & \textbf{0.767} & \textbf{10.28} \\
\hline
SS2 & \(\times\)   & 86.38  & 36.24 & 0.943 & 8.11  \\
SS2 & \checkmark   & \textbf{50.69}  & \textbf{41.36} & \textbf{0.974} & \textbf{3.99}  \\
\hline
\end{tabular}
\end{table}

Fine-tuning markedly improves all metrics for both input types. However, even after fine-tuning, the reconstruction accuracy on synthetic Sentinel-2 data does not fully match that achieved during pre-training on masked hyperspectral data. This is likely because, during pre-training, band-wise masking is applied randomly, allowing adjacent bands to provide context for reconstruction, whereas with synthetic Sentinel-2 data only the spectral ranges corresponding to Sentinel-2 bands are preserved, creating larger continuous spectral gaps to be reconstructed. Furthermore, the comparatively poorer metrics for real Sentinel-2 data suggest additional confounding factors such as cloud occlusion differences, data processing variations, and sensor discrepancies that may complicate the reconstruction from this modality.

\paragraph{Data Efficiency}
We evaluated the impact of training dataset size on reconstruction performance by fine-tuning our model on varying fractions (0.1\%, 1\%, 10\%, and 100\%) of real Sentinel-2 data. Figure~\ref{fig:masked_reconstruction} plots the mean absolute error (MAE) versus the dataset fraction for two scenarios: one with randomly initialized weights and one using pre-trained weights. Results were averaged over multiple runs using different random seeds, with the shaded areas representing the standard deviation. Across all dataset fractions, models initialized with pre-trained weights consistently achieved lower MAE than those trained from scratch, indicating that pre-training effectively captures transferable spectral dependencies and reduces the need for extensive labeled data.

\begin{figure}[!h]
    \centering
    \includegraphics[width=\columnwidth]{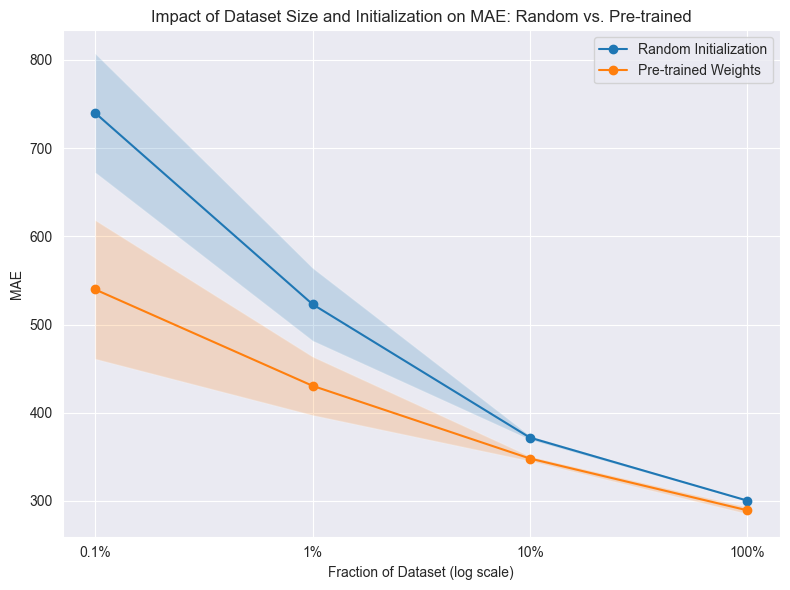}
    \caption{Impact of dataset fraction on reconstruction performance. The plot shows MAE as a function of dataset fraction for models fine-tuned from scratch (random initialization) and with pre-trained weights, with shaded areas indicating standard deviations over multiple runs.}
    \label{fig:masked_reconstruction}
\end{figure}

\begin{table*}[!ht]
\centering
\small
\caption{Performance metrics for CH\textsubscript{4}, NO\textsubscript{2}, and CO\textsubscript{2} detection across different inputs.}
\label{tab:gas_evaluation_combined}
\begin{tabular}{l|l|c|cccc}
\hline
\textbf{Gas} & \textbf{Input} & \textbf{Reconstructed} & \textbf{MAE↓} & \textbf{MSE↓} & \textbf{RMSE↓} & \textbf{R\textsuperscript{2} Score ↑} \\
\hline \hline 
\Tstrut
\multirow{5}{*}{CH\textsubscript{4}} 
& \multirow{2}{*}{Sentinel-2}          & \(\times\)       & \textbf{7.48\(\pm\)0.10}   & 176.24\(\pm\)6.57  & 13.27\(\pm\)0.25  & 0.895\(\pm\)0.004  \\
& & \checkmark      & 7.86\(\pm\)0.07   & \textbf{170.25\(\pm\)8.58}  & \textbf{13.05\(\pm\)0.33}  & \textbf{0.899\(\pm\)0.005}  \\
\cline{2-7}\Tstrut

& \multirow{2}{*}{Synthetic Sentinel-2}  & \(\times\)       & 7.88\(\pm\)0.10   & 192.08\(\pm\)7.80  & 13.86\(\pm\)0.28  & 0.886\(\pm\)0.005  \\
&   & \checkmark    & \textbf{7.58\(\pm\)0.17}   & \textbf{178.43\(\pm\)11.96} & \textbf{13.35\(\pm\)0.44}  & \textbf{0.894\(\pm\)0.007}  \\
\cline{2-7}\Tstrut
& EnMAP  & \(\times\)   & \textbf{7.25\(\pm\)0.11}  & \textbf{139.27\(\pm\)11.05} & \textbf{11.80\(\pm\)0.46} & \textbf{0.917\(\pm\)0.007} \\
\hline \hline
\Tstrut
\multirow{5}{*}{NO\textsubscript{2}} 
& \multirow{2}{*}{Sentinel-2}         & \(\times\)          & \textbf{10.0\(\pm\)0.0\(\times10^{-6}\)}  & \textbf{5.4\(\pm\)0.1\(\times10^{-10}\)}  & \textbf{23.3\(\pm\)0.3\(\times10^{-6}\)}  & \textbf{0.858\(\pm\)0.003} \\
&    & \checkmark   & 10.6\(\pm\)0.1\(\times10^{-6}\)  & 6.1\(\pm\)0.6\(\times10^{-10}\)  & 24.7\(\pm\)1.2\(\times10^{-6}\)  & 0.841\(\pm\)0.016 \\
\cline{2-7}\Tstrut
& \multirow{2}{*}{Synthetic Sentinel-2}         & \(\times\)      & 9.6\(\pm\)0.5\(\times10^{-6}\)   & 4.8\(\pm\)1.5\(\times10^{-10}\)  & 21.7\(\pm\)3.3\(\times10^{-6}\)  & 0.875\(\pm\)0.039 \\
&   & \checkmark   & \textbf{9.4\(\pm\)0.2\(\times10^{-6}\)}   & \textbf{4.3\(\pm\)0.3\(\times10^{-10}\)}  &\textbf{ 20.8\(\pm\)0.7\(\times10^{-6}\)}  & \textbf{0.887\(\pm\)0.007} \\
\cline{2-7}\Tstrut

& EnMAP  & \(\times\)   & \textbf{9.0\(\pm\)0.1\(\times10^{-6}\)}  & \textbf{4.1\(\pm\)0.5\(\times10^{-10}\)}  & \textbf{20.2\(\pm\)1.3\(\times10^{-6}\)}  & \textbf{0.893\(\pm\)0.013} \\
\hline \hline
\Tstrut
\multirow{5}{*}{CO\textsubscript{2}} 
& \multirow{2}{*}{Sentinel-2}       & \(\times\)         & 1.22\(\pm\)0.08  & 3.46\(\pm\)0.49  & 1.86\(\pm\)0.13  & 0.370\(\pm\)0.089 \\
&   & \checkmark    & \textbf{1.03\(\pm\)0.03}  & \textbf{2.09\(\pm\)0.02}  & \textbf{1.45\(\pm\)0.01}  & \textbf{0.620\(\pm\)0.004} \\
\cline{2-7}\Tstrut
& \multirow{2}{*}{Synthetic Sentinel-2}    & \(\times\)            & 1.26\(\pm\)0.01  & 3.61\(\pm\)0.06  & 1.90\(\pm\)0.02  & 0.343\(\pm\)0.011 \\
& \  & \checkmark    & \textbf{1.17\(\pm\)0.01}  & \textbf{3.20\(\pm\)0.07}  & \textbf{1.79\(\pm\)0.02}  & \textbf{0.419\(\pm\)0.012} \\
\cline{2-7}\Tstrut
& EnMAP & \(\times\)    & \textbf{1.20\(\pm\)0.03}  & \textbf{3.38\(\pm\)0.06}  & \textbf{1.84\(\pm\)0.02}  & \textbf{0.385\(\pm\)0.011} \\
\hline \hline
\end{tabular}
\end{table*}

\subsection{Greenhouse Gas Detection Results}
\label{sec:ghg_detection}
Table~\ref{tab:gas_evaluation_combined} presents the performance metrics for greenhouse gas detection using different input types. Across all three gases, models trained on the full original hyperspectral EnMAP data consistently achieve lower errors (MAE, MSE, RMSE) and higher R\(^2\) scores compared to those trained on multispectral inputs (both real Sentinel-2 and synthetic Sentinel-2). These findings demonstrate that the increased spectral coverage and granularity provided by hyperspectral data capture critical absorption features that are valuable for detecting greenhouse gases, affirming the importance of spectral detail in atmospheric monitoring. Furthermore, they establish clear upper and lower baselines for our downstream task. In subsequent evaluations, any improvement over the multispectral baseline when using reconstructed hyperspectral data would indicate that our reconstruction method successfully recovers additional spectral detail beneficial for greenhouse gas detection.

\paragraph{Methane} 
For methane, the reconstructed hyperspectral data improves detection performance across nearly all metrics compared to multispectral inputs. This improvement is observed for both real Sentinel-2 data and synthetic Sentinel-2 data, indicating that the reconstruction method successfully recovers additional spectral detail that enhances methane detection.

\paragraph{Nitrogen Dioxide} 
For NO\(_2\), reconstruction on synthetic Sentinel-2 data moves performance closer to the upper baseline provided by full hyperspectral (EnMAP) data, with improvements seen across all metrics. However, when applied to real Sentinel-2 data, performance degrades in all metrics. A likely explanation is that real Sentinel-2 images may be up to three days offset from the EnMAP capture, while ground-truth NO\(_2\) data is aligned within a one-day window. Given the short atmospheric lifetime of NO\(_2\), this temporal mismatch may significantly impact the results, suggesting that synthetic Sentinel-2 data, with its identical timestamp to the EnMAP data, is more appropriate for NO\(_2\) evaluation.

\paragraph{Carbon Dioxide} 
For CO\(_2\), the reconstructed hyperspectral data shows improvements over multispectral inputs across all metrics. However, these improvements even exceed the performance of the original hyperspectral data, rendering the results inconclusive. A potential explanation is the limited size of the CO\textsubscript{2} dataset, which may make the outcomes more susceptible to random variations.
\section{Conclusion}
\label{sec:conclusion}
This work presents a framework that bridges the gap between hyperspectral and multispectral remote sensing, aiming to combine the broad coverage and high revisit frequency of multispectral platforms with the rich spectral detail of hyperspectral data. Our spectral transformer-masked autoencoder effectively reconstructs synthetic hyperspectral images that preserve valuable spectral information, which can benefit downstream tasks such as greenhouse gas detection, as well as other applications that rely on high spectral detail but are limited by the spatiotemporal constraints of true hyperspectral data.

However, several limitations warrant further investigation. First, the CO\(_2\) dataset used in this study is relatively small, and future work should explore larger, more diverse datasets to better assess the model's performance for CO\(_2\) detection. Second, the advantage of hyperspectral data over multispectral data in the downstream task appears less pronounced for greenhouse gas detection; other downstream tasks—such as mineral identification or vegetation analysis—where spectral detail plays a more critical role may reveal stronger performance gains when using synthetic hyperspectral data. Additionally, the use of Level-2A (surface reflectance) data, dictated by the availability of the EnMAP datasets, may not be ideal for atmospheric applications. Future studies might consider using Top-of-Atmosphere data or focusing on surface-level tasks that are inherently dependent on high spectral resolution.

Despite these challenges, our framework demonstrates promising potential for enhancing the utility of multispectral imagery by recovering hyperspectral-level information. Continued refinement of the reconstruction model and expansion to other datasets and applications will further unlock the benefits of this approach in remote sensing.

{
    \small
    \bibliographystyle{ieeenat_fullname}
    \bibliography{main}
}


\end{document}